[Invited Review]

# Developing ChatGPT for Biology and Medicine: A Complete Review of Biomedical Question Answering

**Qing Li**[1*]**, Lei Li**[1*]**, Yu Li**[1#]

[1]Department of Computer Science and Engineering, Chinese University of Hong Kong, Hong Kong SAR, China

* Co-Author: qingli001@cuhk.edu.hk (Qing Li), leili23@cuhk.edu.hk (Lei Li)

[#] Correspondence: liyu@cse.cuhk.edu.hk (Yu Li)

**Running title**: Developing ChatGPT for Biology and Medicine: A Complete Review of Biomedical Question Answering

**Abstract**

ChatGPT explores a strategic blueprint of question answering (QA) in delivering medical diagnosis, treatment recommendations, and other healthcare support. This is achieved through the increasing incorporation of medical domain data via natural language processing (NLP) and multimodal paradigms. By transitioning the distribution of text, images, videos, and other modalities from the general domain to the medical domain, these techniques have expedited the progress of medical domain question answering (MDQA). They bridge the gap between human natural language and sophisticated medical domain knowledge or expert manual annotations, handling large-scale, diverse, unbalanced, or even unlabeled data analysis scenarios in medical contexts. Central to our focus is the utilizing of language models and multimodal paradigms for medical question answering, aiming to guide the research community in selecting appropriate mechanisms for their specific medical research requirements. Specialized tasks such as unimodal-related question answering, reading comprehension, reasoning, diagnosis, relation extraction, probability modeling, and others, as well as multimodal-related tasks like vision question answering, image caption, cross-modal retrieval, report summarization, and generation, are discussed in detail. Each section delves into the intricate specifics of the respective method under consideration. This paper highlights the structures and advancements of medical domain explorations against general domain methods, emphasizing their applications across different tasks and datasets. It also outlines current challenges and opportunities for future medical domain research, paving the way for continued innovation and application in this rapidly evolving field. This comprehensive review serves not only as an academic resource but also delineates the course for future probes and utilization in the field of medical question answering.

**Key words**

ChatGPT; Medical question answering; Nature language processing; Multimodal paradigms; Large language models

## Introduction

Recently, ChatGPT (Chat Generative Pre-trained Transformer) explores the significant success of general domain question answering (GDQA) and has subsequently permeated from natural language processing (NLP), revolutionizing the way computers understand, interpret, and interact with human natural language, to multimodal paradigms, answering questions that involve multiple modalities, such as text, images, audio, or video. Advancements in pre-trained language representation models such as BERT (Devlin et al, 2018), GPT-2 (Radford et al, 2019), GPT-3 (Brown et al, 2020), ChatGPT (OpenAI, 2022), PaLM (Wei et al, 2022), and LLaMA (Touvron et al, 2022) provide a deeper understanding, reasoning, and generation abilities for general domain question answering, tailored the pre-trained model for a diverse range of tasks. Benefiting from diverse modalities of data, multimodal models such as MCAN (Yu et al, 2019), CLIP (Radford et al, 2021), Flamingo (Alayrac et al, 2022), StableDiffusion (Rombach et al, 2022), GPT-4 (Nori et al, 2023), and MiniGPT-4 (Zhu et al, 2023) offer vision language question answering and comprehensive solution for image processing.

Medical domain question answering (MDQA) has received much attention due to its practical importance in enhancing the deployment of medicine and healthcare, bringing an emerging number of methods built on top of mechanisms of GDQA. The attributes and procedures of GDQA and MDQA are illustrated in Figure 1 (**Figure 1**). Advanced efforts primarily concentrate on biomedical language representation, such as ChestXRayBERT (Cai et al, 2020), PubMedBERT (Gu et al, 2021), BioLinkBERT (Yasunaga et al, 2022), BioGPT (Luo et al, 2023), MedicalGPT (Xu et al, 2023), PMC-LLaMA (Wu et al, 2023), as well as the vision-language model to align biomedical vocabulary, such as MedFuseNet (Sharma et al, 2020), PubMedCLIP (Eslami et al, 2021), Flan-PaLM (Singhal et al, 2022), PaLM-E (Driess et al, 2023), XrayGPT (Thawkar et al, 2023), Med-Flamingo (Moor et al, 2023), AltDiffusion (Ye et al, 2023), CephGPT-4 (Ma et al, 2023). The interrelationship of GDQA and MDQA and their spreading process and timeline of corresponding major models are illustrated in Figure 2 (**Figure 2**).

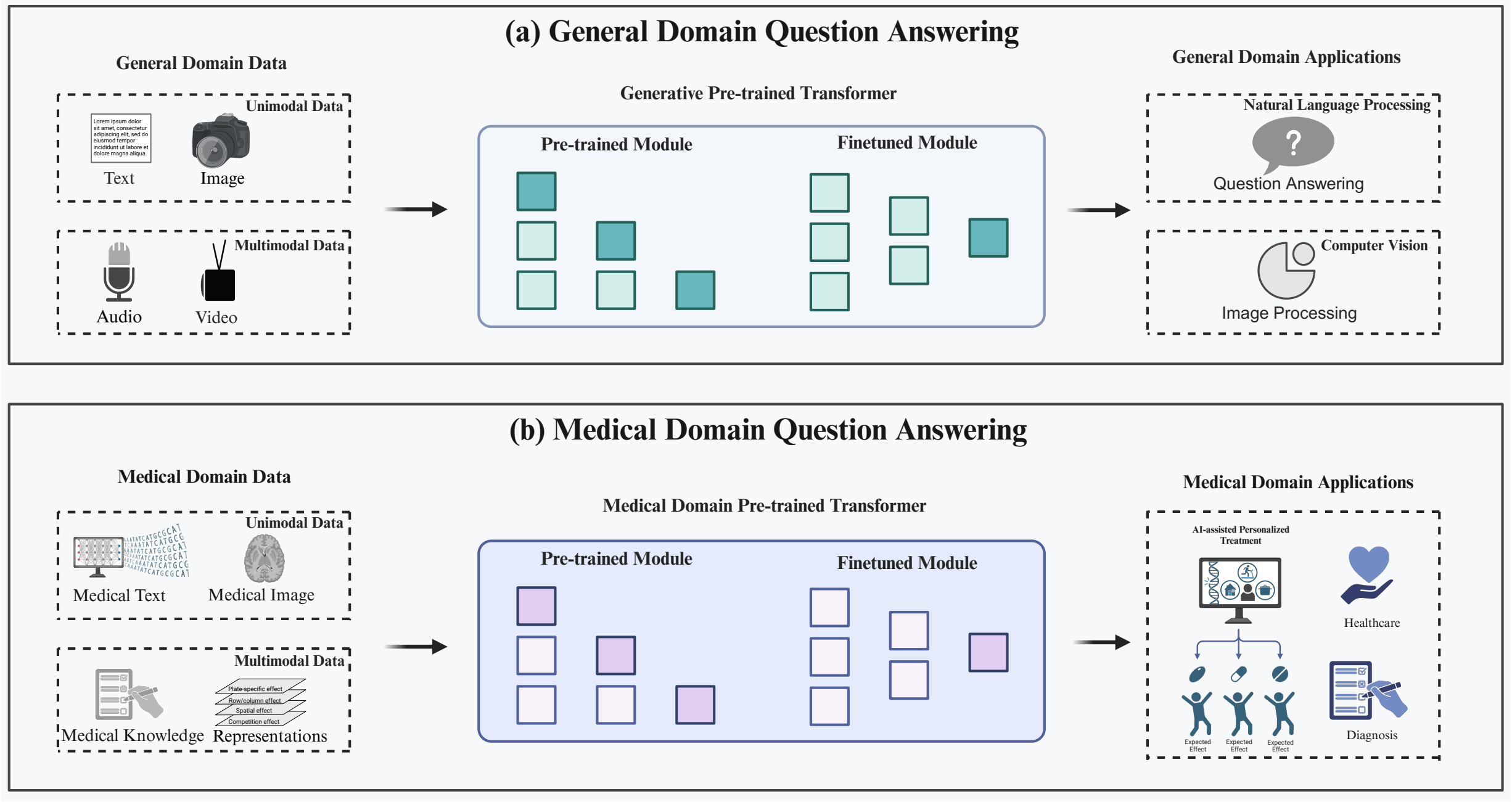

**Figure 1.** General domain question answering (GDQA) and medical domain question answering (MDQA) attributes and procedures. (a) General domain question answering. The top of the figure illustrates the main steps involved in unimodal and multimodal generative pre-trained transformer models for general domain question answering. These models are trained on various types of data such as text, images, audio, and video for unimodal, and trained on multiple data forms like audio and video for multimodal. These models enable general domain question answering applications in areas related to natural language processing and computer vision. (b) Medical domain question answering. The bottom of the figure demonstrates the key steps of unimodal and multimodal MDQA models that are trained on specific types of medical data like medical text and images for unimodal, and trained on various types of medical data, including medical knowledge and multimodal representations of medical data for multimodal. These models enable medical applications such as AI-assisted personalized treatment, healthcare, and diagnosis. It is observed that the structure and mechanisms of MDQA models are primarily developed based on the GDQA generative pre-trained transformer model.

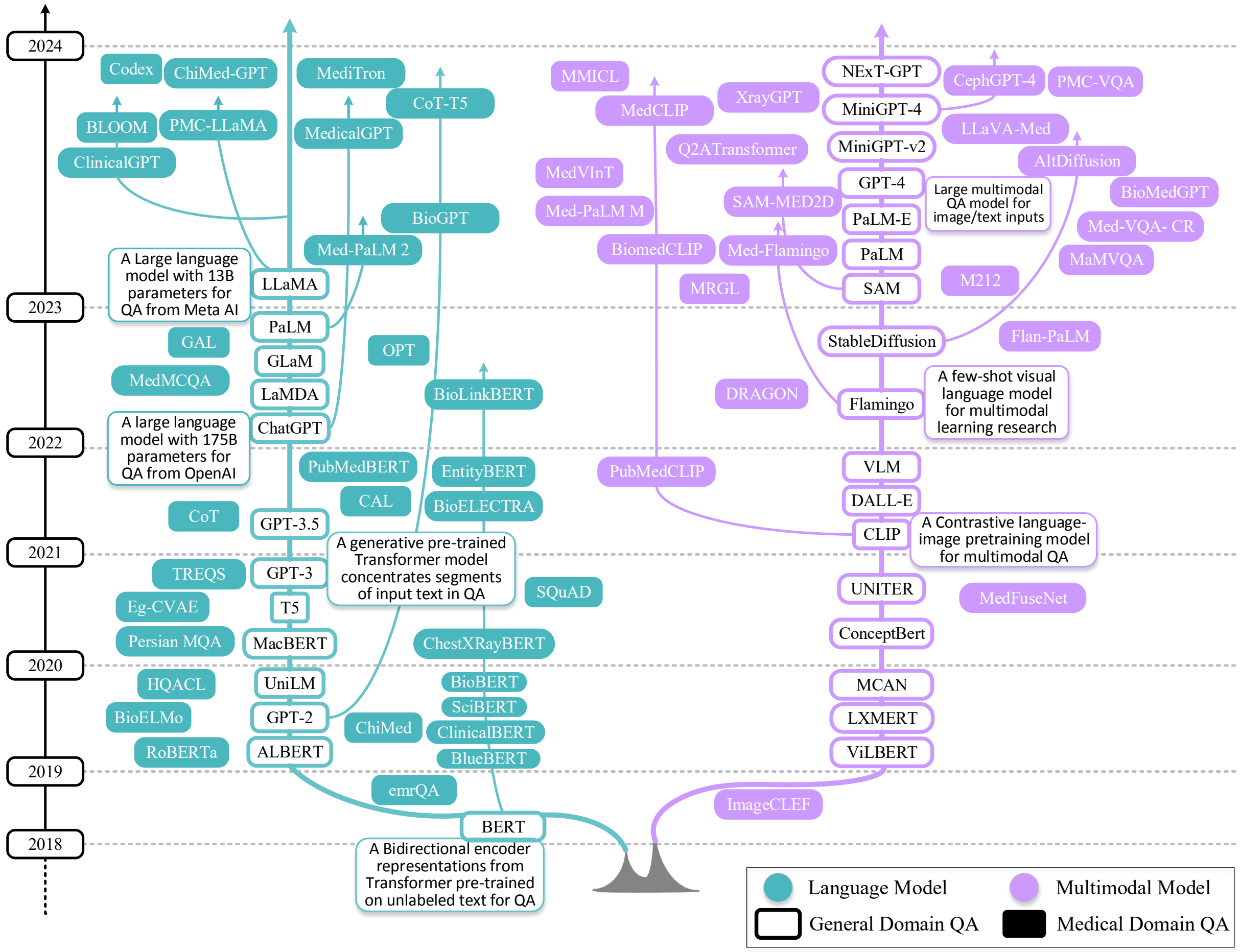

**Figure 2.** General domain and medical domain question answering evolutionary tree. There are general domain QA models represented as hollow rectangles and medical domain QA models represented as solid rectangles, where a part of medical domain QA models are built on top of the general domain QA models which are connected with arrows. Milestones in the general domain include language models: BERT, GPT-3, ChatGPT, and LLaMA, as well as multimodal models: CLIP, StableDiffusion, and GPT-4, are briefly introduced next to the model. Overall, natural language QA in green and multimodal QA in red are two significant branches with explosive growth from 2023 that form the QA evolutionary tree together.

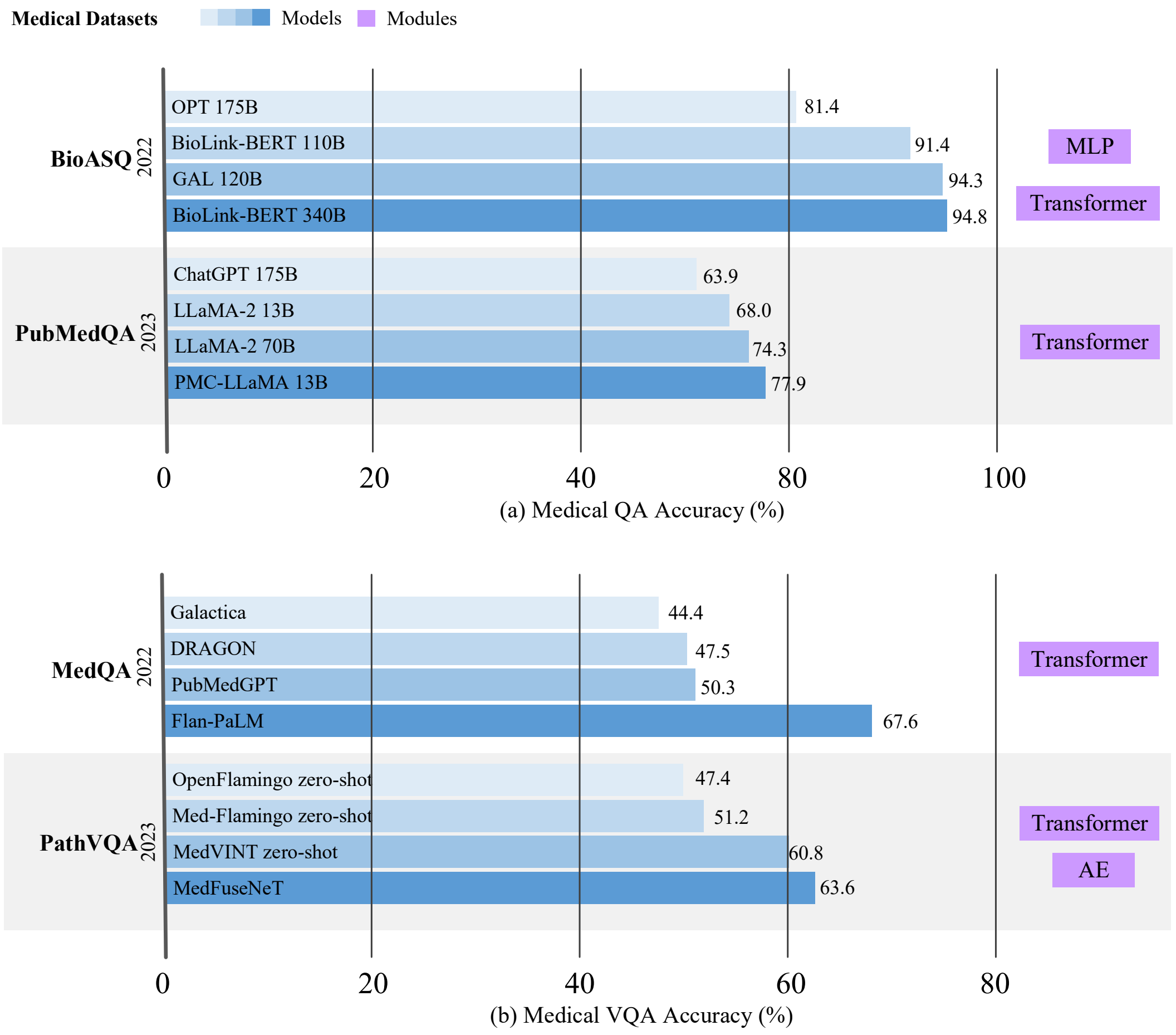

**Figure 3.** Medical question answering (QA) and vision question answering (VQA) model performance examples on popular datasets. (a) Medical QA language model performance examples of the past two years, each year ranked a part of comparable model performance colored from light blue to dark blue with state-of-the-art deep learning modules represented as purple squares. (b) Medical VQA multimodal model performance examples of the past two years, each year ranked a part of comparable models colored from light blue to dark blue with state-of-the-art deep learning modules represented as purple squares. Noticeable, medical data are highlighted in bold, with the horizontal axis denoting the accuracy in percentage.

As a flourish of medical question answering, a succinct review is essential, capturing the avant-garde uses of language models and multimodal paradigms shift from the general domain to the medical domain. These models can be categorized into unimodal medical question answering and multimodal question answering, which are further composed of tasks, mechanisms, and targets in the general domain and medical domain. To decipher their capabilities of answering questions related to medicine and healthcare, we summarize their technical advancements in solving distinct tasks, model structures and utilizations, medical domain datasets, and the corresponding performance in each section. Additionally, the performance of medical domain models on popular medical question answering datasets is illustrated in Figure 3 (**Figure 3**).

The field of language models for question answering has seen the rise of numerous models that have significantly pushed the boundaries in comprehending and encapsulating the intricacies of human language. A summary of medical language question answering models is provided in Table 1 (**Table 1**). These models have been instrumental in a variety of tasks, such as reading comprehension, sentiment analysis, and conversational AI. Their ability to understand context, decipher semantics, and generate human-like responses has revolutionized how we interact with technology, paving the way for more intuitive and natural user experiences. Moreover, multimodal models in medical domain question answering leverage various modalities such as medical images, clinical notes, patient records, and scientific literature to enhance the question answering process. A summary of medical multimodal question answering models is provided in Table 2 (**Table 2**). Both unimodal medical question answering and multimodal medical vision question answering datasets as well as the metric for each dataset are summarized in Table 3 (**Table 3**).

This review is instrumental in bridging the current knowledge gap of ChatGPT and biology, illuminating the structure of medical QA and VQA models, including their contributions, and the challenges and opportunities they pose. In the first chapter of this paper, we will delve into the development of unimodal (language) question-answering models. We will review early classical models, such as rule-based question-answering systems, and how they gradually evolved into modern question-answering models based on deep learning. Furthermore, in the second chapter, we will shift our focus toward multimodal question-answering models, expanding the scope to the fusion of vision and language. We will examine how these models excel in tasks that involve the combination of images and text. It is worth noting that in the latter parts of both Chapter 1 and Chapter 2, we will specifically explore how these models have been extended to the medical domain. The application of question-answering models in the medical field presents unique challenges and opportunities. We will provide a detailed account of the development of question-answering models in the medical domain and highlight potential applications, such as assisting doctors in rapid and accurate diagnosis or providing personalized medical information to patients. In the third chapter, we will introduce an in-depth discussion of the challenges and opportunities in the current field. We will focus on several aspects including model capability and the difficulties faced by models when dealing with specialized terminologies and complex structures in medical texts, the scarcity of data and the uncertainty in the medical domain, and the computational complexity of models when processing large-scale multimodal data, etc. Finally, in the concluding section, we will review the entire research and emphasize the contributions of this paper and its implications for future research. We will propose possible research directions to stimulate further exploration of ChatGPT-related question-answering models in the medical domain within the academic community.

## 1. Unimodal Medical Question Answering

Unimodal models in the medical question answering field demonstrate a commendable proficiency in processing singular data types. These models, by focusing on either textual or visual data, can delve deeply into the nuances of the chosen modality, thereby achieving a high level of specialization. This focused approach allows unimodal models to extract intricate patterns and relationships within the data, leading to robust and accurate predictions. Furthermore, their simplicity relative to multimodal models often results in more efficient training and inference, making them a practical choice for many applications. The unimodal paradigm, with its emphasis on depth over breadth, continues to play a vital role in advancing the field of medical question answering.

### 1.1 Unimodal Models in General Domain Question Answering

General domain question answering with unimodal models mainly advocates for the assembly of a dataset that is as expansive and heterogeneous as feasible, with the aim of amassing demonstrations of tasks in natural language across a broad spectrum of domains and contexts. In natural language processing, they have been applying transfer learning techniques to pre-train language models on extensive, unlabeled corpora from the general domain for question answering.

This includes resources such as Wikipedia articles, web text, book corpora, Gigaword, and web crawls (Common Crawl).

The field of language modeling has seen significant advancements with the introduction of models such as ALBERT (Lan et al, 2020), UniLM (Dong et al, 2019), MacBERT (Cui et al, 2020), T5 (Raffel et al, 2020), LaMDA (Thoppilan et al, 2022), GLaM (Du et al, 2022), and PaLM (Wei et al, 2022). Each of these models brings unique capabilities and improvements to various natural language processing tasks. ALBERT scales with lower memory and increased speed of BERT by using parameter-reduction techniques that share the same weights across its Transformer layers. Despite having a smaller memory footprint, it maintains a computational cost similar to a BERT-like architecture due to the iteration through the same number of repeating layers. It has improved the state-of-the-art significantly for all three benchmarks, achieving notable scores on GLUE, SQuAD 2.0 test F1, and RACE test accuracy. UniLM employs a shared Transformer network to achieve unified modeling. It uses specific self-attention masks to dictate the context that conditions the prediction. UniLM has demonstrated superior performance on five natural language generation datasets, including CNN/DailyMail abstractive summarization, Gigaword abstractive summarization, CoQA generative question answering, SQuAD question generation, and DSTC7 document-grounded dialog response generation.

Instead of using the [MASK] token, which is absent in the fine-tuning stage, MacBERT substitutes the word with a similar one. It enhances RoBERTa in numerous aspects, particularly in its masking strategy that employs MLM as a correction (Mac). Experimental outcomes indicate that MacBERT can attain leading-edge performances on a multitude of NLP tasks. T5 introduces a unified text-to-text format applicable to all tasks. This methodology enables the consistent use of the same model, loss function, and hyperparameters across an extensive array of NLP tasks. The efficacy of T5 is largely contingent on the specific task and the caliber of the fine-tuning. Nevertheless, it has been demonstrated to accomplish top-tier results on numerous NLP tasks. LaMDA is constructed by meticulously fine-tuning a suite of Transformer-based neural language models that are tailored for dialog. These models are educated to utilize external knowledge sources. LaMDA has the capability to access a variety of symbolic text processing systems, which include a database, a real-time clock and calendar, a mathematical calculator, and a system for natural language translation. In every dimension and across all model sizes, LaMDA significantly surpasses the performance of the pre-trained model. Quality metrics, such as Sensibleness, Specificity, and Interestingness, generally see an improvement with the increase in the number of model parameters, irrespective of whether fine-tuning is applied.

Owing to sparsity, GLaM can be trained and served efficiently in terms of computation and energy consumption. The comprehensive version of GLaM encompasses 1.2T total parameters distributed across 64 experts per Mixture of Experts (MoE) layer, with a total of 32 MoE layers. However, during inference, it only activates a subnetwork of 97B parameters, which constitutes 8% of the 1.2T parameters, for each token prediction. When compared to a dense language model, GPT-3 (175B), GLaM's performance is favorable, demonstrating significantly enhanced learning efficiency across 29 public NLP benchmarks spanning seven categories. PaLM, developed by Google Research, is a powerful language model with 540 billion parameters. It was trained using the Pathways system across multiple TPU v4 Pods. It achieved state-of-the-art accuracy on the GSM8K math word problems benchmark and reached 57.8% hardware FLOPs utilization, the highest for large language models at this scale. The training scaled to 6144 chips, the largest TPU-based system configuration used for training to date. These advancements in language modeling continue to push the boundaries of what is possible in natural language processing tasks.

Language modeling has undergone substantial evolution with the advent of models such as GPT-2, LLaMA, GPT-3/3.5, ChatGPT, and InstrucGPT. Each model introduces unique capabilities and enhancements, contributing significantly to a wide range of tasks in natural language processing. These models collectively symbolize the remarkable progress achieved in this domain. GPT-2 ( Radford et al, 2019) showcases a wide array of abilities, including the generation of conditional synthetic text samples of unparalleled quality. This is achieved by priming the model with an input and having it generate an extensive continuation. It surpasses other language models trained on specific domains, such as Wikipedia, news, or books, without the need for these domain-specific training datasets. Human labelers, who were used for training, preferred the fine-tuned models to the base GPT-2 model (zero-shot) 88% and 86% of the time for sentiment and descriptiveness, respectively. The 1.5B model received a "credibility score" of 6.91 out of 10 from people. LLaMA (Touvron et al, 2022) constitutes a suite of foundational language models, with parameters spanning from 7 billion to 65 billion. These models are trained on trillions of tokens, underscoring the feasibility of training cutting-edge models solely on publicly accessible datasets, thereby eliminating the need for proprietary and inaccessible datasets. LLaMA-13B surpasses GPT-3 (175B) on the majority of benchmarks, while LLaMA-65B contends with the top models, Chinchilla-70B and PaLM-540B. All models have been made publicly accessible to the research community.

GPT-3/3.5 (Brown et al, 2020) stands as the most substantial neural network ever constructed, boasting over 175 billion machine learning parameters. It outstrips preceding large language models in both size and proficiency. The model has undergone training via a method known as generative pre-training, where it acquires knowledge from patterns present in the extensive corpus of internet text. GPT-3 has exhibited remarkable efficacy in generating superior-quality text that mirrors content written by humans. It is capable of producing significant quantities of pertinent and sophisticated machine-generated text with minimal input text. The performance of GPT-3 escalates in tandem with the model's size and the volume of data it has been trained on. GPT-3.5 Turbo has undergone fine-tuning to enhance its performance for particular use cases and to operate these bespoke models at scale. This fine-tuning process augments the model's capacity to adhere to instructions more effectively, format responses consistently, and refine the qualitative aspect of the model output. Preliminary tests have indicated that a fine-tuned variant of GPT-3.5 Turbo can equate to, or even surpass, the capabilities of the base GPT-4 on specific narrow tasks. Moreover, fine-tuning with GPT-3.5 Turbo can accommodate 4k tokens, which is double the capacity of previous fine-tuned models.

Non-domain specific Large Language Model (LLM) ChatGPT (OpenAI et al, 2022) has demonstrated capabilities in deductive reasoning, maintaining a chain of thought, and managing long-term dependencies. It excels in addressing long-range dependencies and producing coherent responses that are contextually appropriate. ChatGPT operates without access to external information sources, generating responses based on the abstract relationships between words within the neural network. The model has exhibited understandable reasoning and offered valid clinical insights, bolstering confidence in its trustworthiness and comprehensibility. It has achieved an accuracy of 60%, meeting the passing threshold for the USMLE exams. Remarkably, ChatGPT has performed at or near the passing threshold for all three USMLE exams without any specialized training or reinforcement. Finally, InstrucGPT (Ouyang et al, 2022) signifies a substantial progression beyond preceding GPT models. It has been trained to encapsulate human intentions and align with user directives, thereby enhancing its accuracy and reducing susceptibility to toxic language. This is accomplished through reinforcement learning and fine-tuning predicated on human feedback, a process known as Reinforcement Learning from Human Feedback (RLHF). When juxtaposed with its predecessor, GPT-3, InstructGPT presents several pivotal enhancements. It generates outputs that are more aligned with user directives, thereby enhancing its accuracy and reducing susceptibility to toxic language.

In summary, unimodal question answering models focus on enhancing the speed and versatility of BERT, such as using parameter-reduction techniques and being fine-tuned for various NLP tasks. There are models specifically for Chinese NLP such as MacBERT modifies RoBERTa, and other encoder-decoder models pre-trained on a mix of tasks such as T5. Conversational models LaMDA and GLaM, where LaMDA developed by Google and GLaM using a sparsely activated mixture-of-experts architecture, and large Transformer model PaLM trained with the Pathways system are efficient modules for unimodal question answering. Other large language models pre-trained on a diverse dataset, such as LLaMA aim to construct a smaller model that outperforms larger ones, e.g., GPT-2.

## 1.2 Unimodal Models in Medical Domain Question Answering

It’s important to note that biomedical literature possesses unique concepts and terminologies that are not typically found within the general domain. Medical domain question answering with unimodal models mainly utilizes resources such as MedQA, MedMCQA, PubMed, PubMedQA, ROOTS, MedDialog, MMLU, and ChiMed.

Unimodal medical question answering has witnessed considerable advancements with the introduction of models based on the Bidirectional Encoder Representations from Transformers (BERT). For example, BioBERT (Lee et al, 2019), PubMedBERT (Gu et al, 2020), and BioLinkBERT (Yasunaga et al, 2022), have made notable contributions in the medical domain, demonstrating the potential of BERT-based models in enhancing the performance of medical question answering systems. BioBERT, introduced in 2019, fine-tuned a single BERT to a pre-trained biomedical language representation model. It adapted the word distribution to biomedical corpora in a variety of biomedical text mining tasks. BioBERT significantly outperformed previous state-of-the-art models on biomedical named entity recognition (0.62% F1 score improvement), biomedical relation extraction (2.80% F1 score improvement), and biomedical question answering (12.24% MRR improvement). PubMedBERT proposes a language model specifically designed for biomedical text analysis based on the BERT architecture pre-trained from scratch using abstracts from PubMed and full-text articles from PubMedCentral. This model achieves state-of-the-art performance on many biomedical NLP tasks, and currently holds the top score on the Biomedical Language, which shows good performance in understanding and extracting information from biomedical literature. BioLinkBERT was introduced, which captures document links such as hyperlinks and citation links to include knowledge that spans across multiple documents. It was pretrained by feeding linked documents into the same language model context, besides using a single document as in BERT. On the BLURB score, BioLinkBERT achieved 83.39 for BioLinkBERT-base and 84.30 for BioLinkBERT-large. On PubMedQA, it scored 70.2 for BioLinkBERT-base and 72.2 for BioLinkBERT-large. On BioASQ, it scored 91.4 for BioLinkBERT-base and 94.8 for BioLinkBERT-large. On MedQA-USMLE, it scored 40.0 for BioLinkBERT-base and 44.6 for BioLinkBERT-large.

Generative Pre-trained Transformers (GPT) are a groundbreaking innovation in the field of natural language processing. Developed by OpenAI, GPT models are based on a Transformer-based architecture and follow a two-stage training procedure1. Initially, a language modeling objective is used on unlabeled data to learn the initial parameters of the neural network model. These models, such as GPT-3, have been successfully introduced to understand medical language. BioGPT-Large excels in three end-to-end relation extraction tasks, BC5CDR, KD-DTI, and DDI, outperforming previous models with F1 scores of 44.98%, 38.42%, and 40.76% respectively, and achieving 78.2% accuracy on PubMedQA. Similarly, BioGPT (Luo et al, 2022), a generative transformer language model pre-trained on large-scale biomedical literature, surpasses previous models on most tasks, matching the performance of BioGPT-Large on the aforementioned tasks and accuracy on PubMedQA. In clinical scenarios, ClinicalGPT (Wang et al, 2023) is a language model explicitly designed and optimized for such contexts. It incorporates extensive and diverse real-world data, including medical records, domain-specific knowledge, and multi-round dialogue consultations in the training process. In comparisons against BLOOM-7B and LLAMA-7B, ClinicalGPT emerged victorious in 89.7% and 85.0% of the cases respectively. Catering to the specific needs of Chinese medical text processing is ChiMed-GPT (Tian et al, 2023). Built upon Ziya-13B-v2, it inherits the capability to process extensive context lengths. The context length of CHIMED-GPT is extended to 4,096, guaranteeing its practical value in the medical domain through enhanced context processing capability. These models collectively represent significant advancements in the field of biomedical language processing.

Significant advancements in large language models (LLMs) have recently ushered in a new era in the field of medical question answering. BioELECTRA (Kanakarajan et al, 2021), a biomedical domain language model pre-trained on PubMed and PubMed Central (PMC) full text articles, achieved a new state-of-the-art with an accuracy improvement of 1.39% on MedNLI and 2.98% on the PubMedQA dataset. OPT (Zhang et al, 2022) presented a collection of auto-regressive large language models, with OPT-175B comparable to GPT-3 but requiring only 1/7th the carbon footprint to develop. GAL (Taylor et al, 2022) introduced a set of specialized tokenization for various scientific notations and achieved a state-of-the-art on LaTeX equations, outperforming GPT-3, as well as on PubMedQA and MedMCQA dev. BLOOM (Scao et al, 2022), a 176B-parameter multilingual language model pretrained on ROOTS with multilingual-focused training, shows competitive performance after multitask finetuning. MediTron-70B (Chen et al, 2023) adapted the Llama-2 language model to the medical domain with group-query attention and outperformed Llama-2-70B and Flan-PaLM-7B on downstream medical tasks. Codex 5-shot CoT (Liévin et al, 2023) developed a large language model focused on multiple prompting scenarios and achieved human-level performances on three datasets. CoT-T5-11B (Kim et al, 2023) fine-tuned Flan-T5 on a large amount of rationales, resulting in stronger few-shot learning capabilities on 4 domain-specific tasks and even outperforming ChatGPT. These advancements highlight the rapid progress and potential of large language models in various domains.

The recent introduction of two models, Med-PaLM2 (Singhal et al, 2023) and PMC-LLaMA (Wu et al, 2023), has significantly reshaped the field of medical question answering, enabling the generation of high-quality responses. Med-PaLM2, a large language model, harnesses the power of Google's large language models to provide high-quality answers to medical questions. It achieved an impressive 86.5% accuracy and was the first language model to perform at an "expert" test-taker level on the MedQA dataset of USMLE-style questions, reaching 85% accuracy. Furthermore, it scored a passing 72.3% on the MedMCQA dataset, which comprises Indian AIIMS and NEET medical examination questions. Concurrently, PMC-LLaMA was unveiled as an open-source language model specifically designed for medical applications. It was developed through data-centric knowledge injection and comprehensive fine-tuning for alignment with domain-specific instructions. Despite consisting of only 13 billion parameters, PMC-LLaMA exhibited superior performance on various public medical question-answering benchmarks, even surpassing ChatGPT. These models represent significant strides in the application of language models to the medical domain.

To summary, in the rapidly evolving field of biomedical language models, several models have made significant contributions. BioBERT, BioLinkBERT, and PubMedBERT have leveraged the BERT architecture, pre-training on biomedical corpora to enhance performance on biomedical NLP tasks. Another foundation model is the GPT, by which BioGPT-Large and ClinicalGPT have demonstrated effectiveness in the biomedical domain, and ChiMed-GPT has shown its potential in Chinese medical text understanding. Other approaches, including pretraining on medical domain information, the application of group-query attention, and the use of specialized tokenization for scientific notations, epitomize the forefront of advancements in the realm of biomedical language models.

## 2. Multimodal Medical Question Answering

Both general domain question answering (GDQA) and medical domain question answering (MDQA) benefit from the incorporation of multimodal models. These models leverage the power of multiple modalities to enhance the accuracy, relevance, and comprehensiveness of question answering. In GDQA, multimodal models enable a deeper understanding of diverse topics, while in MDQA, they assist in leveraging medical data for accurate diagnosis, treatment recommendations, and medical research support.

### 2.1 Multimodal Models in General Domain Question Answering

General domain question answering involves answering questions from a broad range of topics and domains. It aims to develop models that can understand and generate human-like responses to questions posed in natural language. Multimodal models play a crucial role in general domain question answering by incorporating multiple modalities such as text, images, videos, and other forms of data to improve the question answering process. In GDQA, multimodal models leverage the complementary information provided by different modalities to enhance the accuracy and relevance of answers. By incorporating images or videos, these models can provide visual context and offer more comprehensive responses. For example, in a question about identifying a species of bird, a multimodal model can analyze both the textual description and a corresponding image to provide a more accurate answer.

Furthermore, multimodal models in GDQA have the potential to tackle complex questions that require cross-modal reasoning. For instance, when answering a question about the historical significance of a famous landmark, a multimodal model can combine textual information from historical documents with images or videos depicting the landmark's architectural features and cultural importance. This integration of modalities enables a deeper understanding of the question and facilitates the generation of more informative responses. Visual language modeling has witnessed the emergence of several models that have made significant advancements in understanding and representing the relationship between vision and language. These models have contributed to various tasks, such as visual question answering, image segmentation, and multimodal understanding.

In the realm of visual question answering, MCAN (Modular Collaborative Attention Network) (Yu et al, 2019) has introduced a joint attention mechanism that effectively captures relevant image features. This enhanced attention mechanism has demonstrated improved performance in the task of answering questions based on visual content. Addressing the correlation between images and language, CLIP (Contrastive Language-Image Pretraining) (Radford et al, 2021) focuses on developing a deep understanding of the relationship between visual and textual information. By combining visual and text encoders, CLIP demonstrates remarkable performance on diverse visual tasks, including image classification and generation. DALL-E (Ramesh et al, 2021) showcases the potential for creative applications in art and design by generating high-quality images from textual descriptions. By combining natural language processing and image synthesis, DALL-E bridges the gap between language and visual creativity.

Going beyond static images, VLM (Video-Language Model) (Xu et al, 2021) introduces a task-agnostic pre-training approach for video and text inputs. This model outperforms task-specific methods in text-video retrieval tasks by capturing the complex interactions between textual and visual information in videos. These advancements in visual language modeling have also extended to the realm of language generation.

Expanding on previous work, ViLBERT (Vision-and-Language BERT) (Lu et al, 2019) has extended the BERT architecture to comprehensively tackle the challenges of vision and language understanding. By combining visual and textual information, ViLBERT achieves state-of-the-art performance across various tasks, showcasing its effectiveness in capturing the intricate relationship between vision and language. LXMERT (Learning Cross-Modality Encoder Representations from Transformers) (Tan et al, 2019) emphasizes sophisticated reasoning abilities and leverages joint representation learning. This comprehensive vision-and-language model excels in image question answering tasks by effectively integrating visual and textual information. Moving beyond image-based tasks, ConceptBert (Gardères et al, 2020) leverages Knowledge Graphs to enhance Visual Question Answering accuracy. By incorporating structured knowledge into the model's understanding of visual and textual information, ConceptBert improves the accuracy and depth of its responses. UNITER (UNiversal Image-TExt Representation) (Chen et al, 2020) presents a joint image-text representation learning model that surpasses previous methods in various vision-and-language tasks. By effectively integrating image and text information, UNITER achieves strong performance across a range of multimodal tasks, demonstrating its versatility and effectiveness. StableDiffusion (Rombach et al, 2022) addresses the computational challenges of generating high-resolution images, achieving promising results while reducing computational demands. In the domain of image segmentation, SAM (Kirillov et al, 2023) introduces the promptable Segment Anything Model architecture along with a large-scale dataset called SA-1B. Through prompt engineering, SAM demonstrates its potential to accurately delineate objects and regions within images. Flamingo demonstrates its versatility and adaptability by enabling zero-shot adaptation to new tasks in both vision and language. Furthermore, PaLM (Pathways Language Model) (Bosma et al, 2023) achieves groundbreaking performance in benchmarks for language comprehension and generation, leveraging its densely activated, Transformer-based architecture. On this basis, PaLM-E (Driess et al, 2023) combines a Vision Transformer with a language model, showcasing improvements in data efficiency and transfer learning across various robot embodiments.

On the language understanding front, GPT-3.5, an enhanced version of GPT-3, addresses limitations in language understanding and generation. With its advanced architecture and the use of prompt engineering and ensemble methods, GPT-3.5 achieves state-of-the-art results on various language tasks and shows promise in medical benchmarks, enabling more accurate and contextually relevant responses. Building upon the successes of its predecessors, GPT-4 (Nori et al, 2023) further improves language generation with an advanced architecture that captures long-range dependencies and excels in understanding multimodal content. This model demonstrates enhanced performance and generates more coherent and contextually relevant text. MiniGPT-v2 (Chen et al, 2023) serves as a unified interface for vision-language tasks, achieving strong performance on visual question answering and grounding. MiniGPT-4 further enhances multimodal capabilities by combining a frozen visual encoder with a large language model. NExT-GPT (Wu et al, 2023) is an end-to-end multimodal language model capable of processing any combination of text, images, videos, and audio. It achieves universal multimodal understanding and supports any-to-any modality input and output.

Overall, these models have made substantial strides in enhancing visual language modeling, effectively tackling various challenges in tasks such as image understanding, language comprehension, and generation within multimodal contexts.

### 2.2 Multimodal Models in Medical Domain Question Answering

Medical domain question answering focuses specifically on answering questions related to medicine and healthcare. It requires models to understand medical terminology, concepts, and domain-specific knowledge to provide accurate and reliable answers. In MDQA, multimodal models can effectively integrate textual and visual information to provide precise answers to medical questions. For example, when asked about the characteristics of a skin lesion, a multimodal model can combine textual descriptions from medical records with corresponding images to offer a more accurate diagnosis. By considering both modalities, the model can identify relevant visual patterns and correlate them with textual information to generate well-informed answers. The integration of multimodal data in MDQA also allows for more sophisticated tasks, such as treatment recommendation or medical research support. By analyzing clinical notes, patient records, and medical images, multimodal models can assist healthcare professionals in making informed decisions. For instance, a multimodal model can analyze a patient's electronic health records, medical images, and relevant scientific literature to recommend personalized treatment options for a specific medical condition.

The field of medical visual question answering (VQA) has experienced the rise of numerous impressive models that integrate computer vision and natural language processing techniques. These models have had a transformative impact on clinical decision support, image interpretation, and patient education by facilitating the fusion of visual information from medical images with textual questions. This paper investigates several noteworthy models that have played a crucial role in advancing medical VQA.

ImageCLEF (Abacha et al, 2019) introduced the Medical Visual Question Answering (VQA-Med) task, which involved creating a dataset comprising radiology images and question-answer pairs. Participating systems demonstrated strong performance, highlighting the potential of medical VQA for clinical decision support, image interpretation, and patient education. MedFuseNet (Sharama et al, 2023), on the other hand, addresses the challenges associated with multimodal inputs in medical VQA. It proposes an attention-based multimodal deep learning model that improves answer generation tasks. MedFuseNet is effective on real-world medical VQA datasets, such as MED-VQA (radiology-based) and PathVQA (pathology-based). Recently, M2I2 (Li et al, 2023) presents a self-supervised framework for medical Visual Question Answering (VQA). By combining different self-supervised methods and utilizing a medical image caption dataset, M2I2 addresses the challenge of limited training data. It achieves improved performance on three public medical VQA datasets: VQA-RAD, PathVQA, and Slake, with absolute accuracy improvements of 1.3%, 13.6%, and 1.1% compared to previous approaches, respectively. In addition, the MedVInT model addresses the challenges of visual-language understanding in the medical field. It combines a pre-trained vision encoder with a large language model and achieves state-of-the-art performance on the newly introduced PMC-VQA dataset. The study emphasizes the significance of this dataset, which surpasses existing ones in size and diversity, enabling comprehensive evaluation and advancements in medical visual question answering.

Transferring successful methods from the general domain to the medical domain has been a focus of research in recent years. By adapting and fine-tuning general domain language models on medical-specific data, researchers aim to create specialized models for medical language understanding and generation. This cross-domain transfer approach harnesses the power of general domain models while addressing the unique challenges of the medical domain. The advancements in this area offer valuable insights into leveraging existing resources to enhance language processing in specialized domains. For instance, MMICL (Zhao et al, 2023) improves vision-language models (VLMs) by addressing their limitations in understanding complex multi-modal prompts. It introduces a new context scheme and the MIC dataset, enabling efficient handling of multi-modal inputs. MMICL achieves state-of-the-art performance on vision-language tasks, especially on challenging benchmarks like MME and MMBench. It effectively handles text-to-image references, relationships between multiple images, and in-context multi-modal demonstrations. MMICL also reduces language bias in VLMs, resulting in impressive performance. Med-Flamingo is an improved version of the Flamingo model that learns from limited examples in real-time. It undergoes a comprehensive clinical evaluation study for medical visual question answering (VQA) and achieves exceptional performance on conventional VQA datasets. Med-Flamingo incorporates pre-training datasets like PMC-OA and MTB, as well as a unique USMLE-style evaluation dataset that combines medical VQA with complex medical reasoning across specialties. It shows promise for clinical applications in the field of medical VQA. Inspired by the widespread utilization of StableDiffusion, AltDiffusion (Ye et al, 2023) is a multilingual Text-to-Image (T2I) diffusion model that supports eighteen languages. It incorporates a multilingual text encoder and a pretrained English-only diffusion model, achieving superior performance in generating high-quality images and capturing culture-specific concepts while maintaining comparable image quality. SAM-Med2D (Cheng et al, 2023) is a specialized version of the Segment Anything Model (SAM) specifically designed for medical image segmentation. It addresses the domain gap between natural and medical images by fine-tuning SAM on a large-scale medical dataset. SAM-Med2D incorporates improved encoder and decoder components and achieves superior performance and generalization on diverse modalities, anatomical structures, and organs, as demonstrated in comprehensive evaluations and validations on multiple datasets from the MICCAI 2023 challenge.

Researchers have been investigating the potential application of CLIP, which has shown promising results in cross-modal generation across various domains, in the medical field. PubMedCLIP focuses on Medical Visual Question Answering (MedVQA) and outperforms the original CLIP and a popular MAML network, achieving improvements of up to 3% in accuracy. BiomedCLIP (Zhang et al, 2023) presents a large-scale study on domain-specific pretraining for biomedical vision-language processing (VLP) and surpasses previous VLP approaches, even outperforming radiology-specific state-of-the-art models. MedCLIP (Wang et al, 2023) addresses the application of visual text contrastive learning to medical images and reports, achieving high data efficiency and superior performance in various tasks. These studies emphasize the advantages of adapting CLIP specifically for the medical domain, enhancing performance and overcoming challenges associated with limited and imbalanced medical training data.

On the other hand, recent studies have emphasized the successful implementation and customization of the PaLM model within the medical field. Flan-PaLM (Singhal et al, 2022) focuses on instruction fine-tuning and demonstrates its effectiveness in scaling the number of tasks, model size, and incorporating chain-of-thought data. The study shows that Flan-PaLM, a 540B-parameter model finetuned on 1.8K tasks, outperforms previous models like PaLM on multiple evaluation benchmarks, achieving state-of-the-art performance. Additionally, Flan-T5 checkpoints exhibit strong few-shot performance compared to larger models. Med-PaLM 2 (Singhal et al, 2023) addresses limitations in medical question-answering by combining improvements in base language models, medical domain fine-tuning, and novel prompting strategies. Med-PaLM 2 achieves promising results on various medical datasets and demonstrates improved accuracy compared to its predecessor. It also receives positive evaluations from physicians and introduces new adversarial question datasets. Moreover, Med-PaLM M (Tu et al, 2023) focuses on a generalist biomedical AI system and presents the MultiMedBench benchmark, demonstrating the potential of generalist models for various medical tasks. It addresses shortcomings in biomedical AI and showcases promising results in generating chest X-ray reports and zero-shot multimodal medical reasoning. The study emphasizes the importance of large-scale biomedical data access, real-world performance validation, and safety considerations.

In parallel with the successful application of transformer-based models and GPTs in general domain visual question answering (VQA), the field of healthcare has witnessed a flourishing development, giving rise to a plethora of medical VQA models. Q2ATransformer (Liu et al, 2023) is a framework that employs a transformer decoder to effectively handle image-question features and candidate answer embeddings. It incorporates answer semantics and reduces the search space for answers, resulting in improved performance on medical VQA tasks. MaMVQA (Manmadhan et al, 2023) tackles the challenges of applying general VQA models to the medical domain by leveraging parallel multi-head attention and a novel semantic term-weighting scheme. It utilizes unsupervised learning for image featurization and domain-specific embeddings to interpret complex clinical terminologies in questions. MaMVQA achieves high accuracy on the VQA-RAD dataset, particularly for open-ended questions. XrayGPT (Thawkar et al, 2023) focuses on chest radiograph interpretation and combines a visual encoder with a language model. It aligns high-quality summaries with X-ray images through interactive conversations, leading to exceptional performance on the MIMIC-CXR dataset. The model benefits from fine-tuning on real patient-doctor conversations and radiology dialogues. BioMedGPT(Luo et al, 2023) addresses the limitations of general-purpose language models in handling biomedical domain questions. It introduces BioMedGPT-10B, a large-scale generative language model that unifies the feature space of molecules, proteins, and natural language. This model demonstrates superior performance on biomedical question-answering tasks and shows promise in accelerating drug discovery and identifying therapeutic targets. CephGPT-4 (Ma et al, 2023) is a multimodal cephalometric analysis and diagnostic dialogue model for orthodontic medicine. It incorporates a fine-tuned version of MiniGPT-4 and VisualGLM, trained on a dataset of cephalometric images and doctor-patient dialogues. The model excels in automatic analysis of cephalometric landmarks and generation of diagnostic reports, addressing challenges in manual landmark annotation and accurate interpretation of cephalometric analysis results.

As the field of medical visual question answering (MedVQA) continues to evolve, several promising research directions are emerging. These directions aim to tackle open-ended problems and leverage knowledge graph-based representation learning to enhance the capabilities of VQA models. By exploring these areas, researchers can unlock new possibilities in understanding medical images and providing accurate and insightful answers to complex visual questions. As an illustration, LLaVA-Med (Li et al, 2023) is a cost-efficient method designed to train a vision-language conversational assistant capable of addressing open-ended research questions related to biomedical images. It highlights the limitations of current general-domain visual assistants in biomedical scenarios and proposes a solution using a comprehensive biomedical figure-caption dataset. The paper's key contribution is the application of curriculum learning to fine-tune a large general-domain vision-language model. Evaluation on a new dataset of 193 questions demonstrates outstanding performance, including a 0.68 F1 score for open-set questions and a 0.91 accuracy score for close-set questions. In addition, Med-VQA-CR (Zhan et al, 2020) introduces a conditional reasoning framework, combining problem-conditional reasoning (QCR) and task-conditional reasoning (TCR), to effectively address open- and closed-ended problems. Integration of QCR and TCR enhances overall accuracy from 66.1% to 71.6%, with open accuracy improving from 49.2% to 60.0%, and closed accuracy from 77.2% to 79.3%, showcasing superior performance on the VQA-RAD dataset and excelling in higher-level reasoning skills, particularly in open-ended tasks. On the other hand, DRAGON (Yasunaga et al, 2022) is a self-supervised method introduced for pretraining a language-knowledge model from text and knowledge graphs (KGs). The authors propose a deeply bidirectional model that effectively combines information from both modalities, demonstrating superior performance over existing models in general and biomedical domains. Evaluation on downstream tasks, including question answering and BioNLP, shows an average improvement of 5% and 8% across different tasks. Besides, MRGL tackles the challenge of medical visual question answering (VQA) by integrating multi-modal relationship graph learning. It addresses the shortcomings of existing medical VQA methods by introducing a comprehensive dataset centered on chest X-ray images, capturing detailed questions related to disease names, locations, levels, and types. The proposed baseline method constructs spatial, semantic, and implicit relationship graphs on image regions, questions, and semantic labels, allowing the model to learn answer reasoning paths for diverse questions.

Overall, In the realm of multimodal visual-language models within the medical domain, a comprehensive exploration of prevalent models reveals their remarkable efficacy in amalgamating medical imaging and natural language processing. These models not only elevate the precision of medical image analysis but also pave the way for a novel approach to holistic comprehension of medical information. In summary, multimodal visual-language models present unparalleled collaborative benefits in the field of medicine, heralding fresh opportunities for advancing medical research and clinical practices.

## 3. Challenges and Opportunities

**Advancements in Information Fusion and Semantic Comprehension.** The complexity and diversity of semantic understanding in general domain questions and answers demand sophisticated models with robust semantic comprehension capabilities. QA models need to be able to understand the nuances and context of questions, as well as generate accurate and detailed answers. This requires not only a deep understanding of the language but also the ability to reason and integrate information from multiple sources. In the expansive field of general domain research on multimodal question-answering (QA) models, the integration of diverse data types poses a central challenge. The heterogeneity of multimodal data, encompassing text, images, and audio, requires innovative approaches to effectively understand and fuse this information. For example, in the context of image and text data, models need to develop techniques for extracting relevant information from images and aligning it with the corresponding textual content. This requires advanced computer vision algorithms and natural language processing techniques to process and extract meaningful information from the different modalities. Furthermore, the complexity and diversity of semantic understanding in general domain questions and answers demand sophisticated models with robust semantic comprehension capabilities. QA models need to be able to understand the nuances and context of questions, as well as generate accurate and detailed answers. This requires not only a deep understanding of the language but also the ability to reason and integrate information from multiple modalities. Achieving this level of semantic comprehension often involves leveraging pre-trained language models, such as transformer-based architectures, and adapting them to handle multimodal inputs. Despite these challenges, the prospect of leveraging multiple sources for information fusion, employing transfer learning to enhance adaptability, and refining real-time feedback systems opens exciting opportunities for the development of intelligent assistants and advanced QA systems. By integrating different data types and improving the models' ability to understand and reason across modalities, multimodal QA models have the potential to greatly enhance user interactions and provide more comprehensive and accurate answers to complex queries.

**Data complexity, uncertainty and multimodal processing.** Texts in the medical field often contain a large amount of professional terms and complex structures, which require sufficient medical knowledge and contextual understanding. In addition, medical data are scarce and full of uncertainty, making it difficult to obtain large-scale and high-quality annotated data. Multimodal question answering models face many opportunities and challenges in the medical field. Texts in the medical field often contain a large amount of professional terms and complex structures, which is a challenge for models because they require sufficient medical knowledge and contextual understanding. In addition, data in the medical field are scarce and full of uncertainty. It is difficult to obtain large-scale and high-quality annotated data, and medical decisions and diagnostic results are often uncertain. These factors bring challenges to the training and inference of multi-modal question answering models. In addition, data in the medical field are usually multi-modal, including text, images, videos, etc. Processing large-scale multi-modal data sets requires a large amount of computing resources and storage space. At the same time, it is necessary to process the alignment and alignment between different modalities. Fusion. Despite these challenges, there are many opportunities for multimodal question answering models in the medical field. They can be used to automate the generation and analysis of medical reports, assist doctors in making clinical decisions, or help patients obtain medical knowledge and answer questions. With the accumulation of medical data and improvement of models, multi-modal question answering models are expected to play a greater role in the medical field.

**Navigating Complexity, Ensuring Privacy, and Advancing Precision Medicine.** The intricacies of medical knowledge pose a significant challenge. Navigating complex terminologies and understanding the nuances of medical information are essential. In the specialized field of medical multimodal question-answering research, the intricacies of medical knowledge pose a significant challenge. Navigating complex terminologies and integrating diverse data types, such as images and texts, demands models capable of comprehending the nuances of medical information. Medical data, encompassing structured and unstructured formats like electronic health records, medical images, clinical notes, research articles, and patient-generated data, requires QA models to process and interpret this varied information for meaningful insights and answers. The development and deployment of multimodal QA models in the medical domain are further complicated by the strict adherence to privacy regulations and ethical standards. Patient data sensitivity necessitates healthcare providers and researchers to safeguard patient privacy and comply with applicable regulations. Designing models and systems that handle sensitive data securely and incorporate privacy-preserving techniques becomes imperative in this context. Amidst these challenges, the medical field presents promising opportunities for advancing precision medicine through accurate diagnostics and treatment recommendations. Multimodal QA models can empower healthcare professionals by offering comprehensive and timely information from diverse sources. They play a crucial role in tasks such as analyzing medical images, extracting relevant information from medical literature, and aiding in the interpretation of clinical data. By automating these functions and enhancing the capabilities of healthcare professionals, multimodal QA models have the potential to revolutionize healthcare practices and enhance patient outcomes. Furthermore, the significance of multimodal QA models in the medical domain is underscored by their ability to facilitate interdisciplinary research. By assisting in the comprehension and utilization of diverse medical data types, such as genetics, imaging, clinical trials, and patient records, these models contribute to accelerating discoveries, promoting collaboration across research domains, and ultimately advancing medical knowledge and patient care.

## 4. Conclusions

To conclude, the sophistication and efficacy of ChatGPT-related language and multimodal mechanisms in addressing medical question-answering challenges, encompassing both unimodal and multimodal dialogues, are noteworthy. These models also tackle other issues such as medical diagnosis, reasoning, machine translation, and image segmentation, and their performance is critically evaluated against conventional approaches. The discussion concludes with an exploration of the challenges and prospects in the realms of medical information, question comprehension, and model capacity to provide an efficient approach to preserving the benefits of the training process while enhancing the resolution of intricate medical dilemmas.

**Acknowledgments**

Qing Li expresses her deep appreciation for the talented collaborators whose invaluable contributions have enabled successful explorations within her laboratory. We apologize to the scientists who contributed to the field but have not been cited due to space limitations. This work was partially supported by a grant from the Research Grants Council of the Hong Kong Special Administrative Region, China [Project No.: CUHK 24204023], and a grant from Innovation and Technology Commission of the Hong Kong Special Administrative Region, China [Project No.: GHP/065/21SZ].

Table 1: A Summary of Medical Language Question Answering Models. State-of-the-art medical question-answering models are summarized as well as their dataset, tasks, technical advancement and their publication details.

| Model Name | Dataset | Tasks | Technical Advancement | Author Name, Publication Year |
|---|---|---|---|---|
| BioBERT | BioASQ | QA, NER, RE | A pre-trained biomedical language representation model that fine-tuned a single Bidirectional Encoder Representation from Transformers (BERT) model | Lee et al., 2019 |
| BioELECTRA | PubMed, PMC | QA, NER, PICOE, RE, SS, DC | A biomedical domain language model pretrained on PubMed and PubMed Central (PMC) full text articles from scratch | Kanakarajan et al., 2021 |
| PubMedBERT | PubMed, PubMed Central | QA, NER, PICOE, RE, SS, DC | A language model for biomedical text analysis based on the BERT architecture pretrained from scratch using abstracts from PubMed and full-text articles from PubMedCentral | Gu et al., 2020 |
| BioLinkBERT | Wikipedia, PubMed | QA, NER | A language model captures document links such as hyperlinks and citation links to include knowledge that spans across multiple documents | Yasunaga et al., 2022 |
| OPT | RoBERTa, Pile, PushShift.io | QA | A collection of auto-regressive large language models ranging from 125M to 175B parameters | Zhang et al., 2022 |
| GAL | PubMedQA, MedMCQA, Galactica Corpus | QA, Reasoning, CP, BU | A large language model uses a set of specialized tokenization for citations, step-by-step reasoning, mathematics, numbers, SMILES formula, amino acid sequences, DNA sequences, etc. | Taylor et al., 2022 |
| ClinicalGPT | cMedQA2, cMedQA-KG, MD-EHR, MEDQA-MCMLE, MedDialog | QA, Diagnosis | A language model explicitly designed and optimized for clinical scenarios by incorporating extensive and diverse real-world data, such as medical records, domain- specific knowledge, and multi-round dialogue consultations in the training process | Wang et al., 2023 |
| BLOOM | GitHub, OSCAR, Common Crawl | PM, MT | A 176B-parameter multilingual language model pretrained on ROOTS with multilingual-focused training | Scao et al., 2022 |
| CoT-T5 | CoT Collection, Flan Collection | Reasoning | A language model that uses Flan-T5 as a base model fine-tuned on a large amount of rationales | Kim et al., 2023 |
| MediTron | MedQA, PubMedQA, MedMCQA, | Reasoning | A large language model adapts the Llama-2 language model to the medical domain with group-query attention through continued pretraining on medical corpus | Chen et al., 2023 |
| Codex 5-shot CoT | USMLE, MedMCQA, PubMedQA | QA, RC | A large language model focus on multiple prompting scenarios including Chain-of-Thought, few-shot, and retrieval augmentation | Liévin et al., 2023 |
| Med-PaLM2 | MedQA, PubMedQA, MedMCQA, MMLU | QA | Med-PaLM 2 is a large language model designed to provide high-quality answers to medical questions that harnesses the power of Google's large language models | Singhal et al., 2023 |
| PMC-LLaMA | MedQA, MedMCQA, PubMedQA, S2ORC, Medical Textbooks | QA, Reasoning | PMC-LLaMA is an open-source language model specifically designed for medical applications through data-centric knowledge injection and comprehensive fine-tuning for alignment with domain-specific instructions | Wu et al., 2023 |
| ChiMed-GPT | CMD;ChiMed, MC, MedDialog | QA, IE | A language model for Chinese medical text processing, which is built upon Ziya-13B-v2 and inherited its capability to pro-cess extensive context lengths | Tian et al., 2023 |
| BioGPT-Large | BC5CDR; KD-DDI, DDI;PubMedQA, HoC | QA, RE, DC | A generative Transformer language model pre-trained on large scale biomedical literature | Luo et al., 2023 |
| BioGPT | PubMed | QA, RE, DC | A domain-specific generative Transformer language model pre-trained on large-scale biomedical literature | Luo et al., 2023 |

Dataset: BC5CDR | BioASO | cMedQA-KG | cMedQA2 | ChiMed | CMD | CoT Collection | DiaBLa | DDI | Flores-101 | Flan Collection | HoC | KD_DDI | MMLU | MD-HER | MedDialog | MedMCQA | Medical Textbooks | MedQA | MedQA-USMILE | PMC | PubMed | PubMed Central | PubMedQA | PushShift.io | Pile | RoBERTa | ROOTS | S2ORC | USMILE | Wikipedia | WikiLingua | WMR14

Task: Biological Understanding, BU | Citation Prediction, CP | Document Classification, DC | Information/Relation Extraction, IR/E | Machine Translation, MT | Name Entity Recognition, NER | PICO (Participants, Interventions, Comparisons and Outcomes entities) extraction, PICOE | Probability Modeling, PM | Question Answering, QA | Reading Comprehension, RC | Reasoning/Diagnosis | Sentence Similarity, SS

Table 2: A Summary of Medical Multimodal Question Answering Models. State-of-the-art medical multimodal question-answering models are summarized as well as their dataset, tasks, technical advancement and their publication details.

| **Model Name** | **Dataset** | **Tasks** | **Technical Advancement** | **Author Name, Publication Year** |
|---|---|---|---|---|
| PubMedCLIP | PubMed | VQA | A contrastive pre-trained model for medical image classification tasks that trained on the ROCO dataset as part of a medical visual question answering model | Eslami et al., 2021 |
| MaMVQA | VQA-RAD | VQA | A multimodal large generative model using the unsupervised Denoising Auto-Encoder (DAE) and language feature based on the concept of mutual information (MI) between the word and the corresponding class label | Manmadhan et al., 2022 |
| MRGL* | ImageCLEF, VQA-RAD | VQA | A multimodal model for medical visual question answering where multi-modal features are implicitly fused by using the multi-head self-attention mechanism | Hu et al., 2023 |
| BiomedCLIP | PMC-15M | VQA, CMR, IC | A contrastive pre-trained model uses PubMedBERT as the text encoder and Vision Transformer as the image encoder | Zhang et al., 2023 |
| PaLM-E | PaLM-E | VQA | A new generalist robotics model that transfers knowledge from varied visual and language domains to a robotics system by transforming sensor data, e.g., images, into a representation through a procedure that is comparable to how words of natural language are processed by a language model | Driess et al., 2023 |
| PMC-VQA | PMC-VQA, VQA-RAD, SLAKE | VQA | A generative-based model for medical visual understanding by aligning visual information from a pre-trained vision encoder with a large language model | Zhang et al., 2023 |
| LLaVA-Med | PubMed Central, VQA-RAD, SLAKE, Path-VQA | VQA | A large vision-language model to align biomedical vocabulary using the figure-caption pairs that learns to master open-ended conversational semantics using GPT-4 generated instruction-following data | Li et al., 2023 |
| Q2ATransformer | VQA-RAD, Path-VQA | VQA | A medical VQA model, integrating the advantages of both classification and generation approaches and provides a unified treatment for close-end and open-end questions | Liu et al., 2023 |
| XrayGPT | MIMIC-CXR, OpenI | VQA | A pre-trained medical vision and language model aligns both a medical visual encoder (MedClip) with a fine-tuned large language model (Vicuna), using a simple linear transformation | Thawkar et al., 2023 |
| Med-PaLM M | MedQA,MedMCQA,PubMedQA,MIMIC, VQA-RAD, Slake-VQA, Path-VQA, VinDr-Mammo, CBIS-DDSM, PrecisionFDA | VQA, RS, RG, IC | A large multimodal generative model that flexibly encodes and interprets biomedical data including clinical language, imaging, and genomics with the same set of model weights | Tu et al., 2023 |
| CephGPT-4 | MD-QA, OCIMM, PMC-VQA, VQA-RAD, SLAKE | VQA | A large-scale multimodal language models based on Minigpt-4, which can automatically analyze cephalometric medical images and provide diagnostic results and treatment advice | Ma et al., 2023 |
| Med-Flamingo | MTB, PMC-OA | VQA, RG | A multimodal few-shot model based on the OpenFlamingo-9B V1 model which uses the CLIP ViT-L/14 vision encoder and the Llama-7B language model as frozen backbones | Moor et al., 2023 |
| BioMedGPT-10B | PubChemQA, UniProtQA | VQA | A multi-modal foundation model in the biomedical domain leverages the incremental fine-tuning of large language models to comprehend biomedical documents and aligns the feature spaces of molecules, proteins, and natural language. | Luo et al., 2023 |
| AltDiffusion | LAION 5B, LAION, Aesthetics | IG | A novel multilingual (Chinese to English) text-to-image diffusion model built on Stable Diffusion that supports eighteen different languages trained by two stages: concept alignment and quality improvement | Ye et al., 2023 |
| SAM-Med2D | 4.6M images and 19.7M masks from public and private datasets | VQA, IS | A medical image segmentation through more comprehensive prompts involving bounding boxes, points, and masks. | Cheng et al., 2023 |
| M2I2 | ImageNet, VQA-RAD, Path-VQA, SLAKE | VQA | A multimodal model pretrained on medical image caption dataset, and finetuned to downstream medical VQA tasks | Li et al., 2023 |
| MMICL | MIC | VQA | A new approach to allow the vision-language model to deal with multi-modal inputs efficiently by a novel context scheme | Zhao et al., 2023 |
| MedFuseNet | Path-VQA, DAQUAR, VQA,VQA 2.0, CLEVR | VQA | An attention based multimodal deep learning model which learns representations by optimal fusion of the multimodal inputs using attention mechanism | Sharma et al., 2021 |
| Med-VQA | VQA-RAD | VQA | A conditional reasoning framework with task-specific reasoning ability realized by the use of an attention mechanism conditioned by task information to guide the importance weighting of multimodal fusion features | Zhan et al., 2020 |
| MedCLIP | MIMIC-CXR, CheXpert; MIMIC, COVID, RSNA | VQA, ZSC, ITR | A contrastive learning model that scales the usable training data in a combinatorial magnitude with low cost by decoupling images and texts | Wang et al., 2022 |
| GPT-4V | Nocaps,Flickr30K,VQA v2, OKVQA,GQA, ScienceQA,VizWiz, OCR_VQA | VQA, IC | A multimodal LLM builds on the work done for GPT-4 and dives deeper into the evaluations, preparation, and mitigation for medical licensing examination questions with images | Wu et al., 2023 |

Dataset: CBIS-DDSM | CLEVR | COVID | CheXpert | DAQUAR | DAQUAR | Flickr30K | GQA | ImageCLEF | ImageNet | LAION | MD-QA | MIC | MIC | MIMIC | MIMIC-CXR | MTB | MedMCQA | Nocaps | OCIMM | OKVQA | OpenI | PMC-15M | PMC-OA | PMC-VQA | PaLM-E | Path-VQA | PrecisionFDA | PubChemQA | PubChemQA | PubMed Central | PubMedQA | ROCO | RSNA | SLAKE | Slake-VQA | ScienceQA | UniProtQA | UniProtQA | VQA | VQA 2.0 | VQA-RAD | VQAv2 | VinDrMammo

Task: Cross-Modal Retrieval, CMR | Image Caption/Classification/Generation/Segmentation, IC/Cl/G/S | Report Summarization/Generation, RS/G | Visual Question Answering, VQA
MMRGL*: Multi-Modal Relationship Graph Learning

Table 3: Medical Question Answering Dataset Overview. Summarizing widely used datasets for language and multimodal question answering in the medical domain, along with the corresponding evaluation metrics. Consult Table 1 and Table 2 for more comprehensive information regarding task-specific datasets evaluated for different models.

| Task Type | Dataset | Data Size (Train/Val/Test) | Metrics | Dataset Description |
|---|---|---|---|---|
| Language Question Answering | MedQA | 61,097 (80%/10%10%) | Accuracy | A medical question bank for professional board exams covering English, simplified Chinese, and traditional Chinese |
| | MedMCQA | 193,155 (182,822, 6,150, 4,183) | Accuracy | Address real-world medical entrance exam questions with deep language understanding |
| | PubMedQA | 273.5K (1k expert-annotated, 61.2k unlabeled and 211.3k artificially generated ) | Accuracy, macro-F1, F1 | Aimed at biomedical research question answering using yes/no/maybe |
| | Flowers102 | 8,189 images (1,030/1,030/6,129) | Recognition rate | Flowers contain 102 types, each class consists of between 40 and 258 images |
| | CUB-200-2011 | 200 categories, 11,788 images | Accuracy | Consisting of 200 bird species with over 11,000 images for fine-grained classification tasks |
| Multimodal Question Answering | PMC-15M | 15,282,336 figure-caption pairs (13.9M/13.6K/726K) | R@k (recall score among top k results) | Given textual description (caption), retrieve the corresponding image, or vice versa |
| | Visual Genome | 108K images, 1773K QApairs | - | Linking language and vision through crowdsourced dense image annotations |
| | VQA 2.0 | 204K images, 614K QApairs | VQA score | Addressing language biases in VQA and emphasizing image understanding |
| | OK-VQA | 14,031 images, 14,055 QApairs | VQA score | Knowledge-based VQA which requires reasoning on External knowledge |
| | VQA-Med-2018 | 2866 images, 6,413 QApairs (5413, 2278/500, 324/500, 264) | BLEU, WBSS, CBSS | Focus on VQA in the medical domain |
| | VQA-RAD | 315 images, 3,515 QApairs (42% (637) open-ended answer types and 58% (878) close-ended) | Simple Acc, Mean Acc, BLEU | VQA dataset with clinicians' 11-type questions and reference answers about radiology images |
| | VQA-Med-2019 | 4200 images, 15292 QApairs (3200, 12792/500, 2000/500, 500) | BLEU, Accuracy | Using question patterns from medical students to create clinically relevant questions in four categories |
| | RadVisDial (Silver-standard) | 91,060 images, (77205, 7340, 6515), 455300 QApairs | F1 score, Macro F1 | Introducing the silver-standard datasets for VQA in radiology, specifically utilizing chest X-rays |
| | RadVisDial (Gold-standard) | 100 images, 500 QApairs | F1 score, Macro F1 | Gold standard dataset comprising dialogues between two expert radiologists discussing specific chest X-rays |
| | PathVQA | 4998 images, 32795 QApairs (16,466 open-ended, other close-ended) (3021, 19755/987, 6279/990, 6761) | Accuracy, BLEU, F1 | The first dataset for pathology VQA |
| | VQA-Med-2021 | VQA: 5500 images, 5500 QApairs (4500, 4500/500, 500/500, 500) VQG: (-/85, 200/100, 302) | Accuracy, BLEU | Including tasks on both Visual Question Answering (VQA) and Visual Question Generation (VQG) |
| | SLAKE | 642 images, 14K QApairs (450, 9849/96, 2109/96, 2070) | Accuracy | A semantically-labeled knowledge-enhanced dataset with an extendable knowledge base for Med-VQA |